\documentclass[twocolumn]{article}
\usepackage[dvipdfmx]{graphicx}
\usepackage{url}
\usepackage{amsmath}
\usepackage{amssymb}
\usepackage{amsfonts}
\usepackage{color}
\usepackage[utf8]{inputenc}
\usepackage[T1]{fontenc}
\usepackage{float}
\usepackage{algorithm}
\usepackage{algorithmic}
\usepackage{subcaption}
\usepackage{caption}
\usepackage{booktabs}
\usepackage{geometry}
\usepackage{authblk}

\title{FedAvg-Based CTMC Hazard Model\\for Federated Bridge Deterioration Assessment}

\author[1]{Takato Yasuno}
\affil[1]{Yachiyo Engineering Co., Ltd., Tokyo, Japan}
\affil[ ]{Email: tkt-yasuno@yachiyo-eng.co.jp}

\date{}

\begin{document}
\maketitle

\begin{abstract}
Bridge periodic inspection records contain sensitive information about public infrastructure,
making cross-organizational data sharing impractical under existing data governance constraints.
We propose a federated framework for estimating a Continuous-Time Markov Chain (CTMC) hazard
model of bridge deterioration, enabling municipalities to collaboratively train a shared benchmark
model without transferring raw inspection records.
Each User holds local inspection data and trains a log-linear hazard model over three
deterioration-direction transitions---Good$\to$Minor, Good$\to$Severe, and Minor$\to$Severe---with
covariates for bridge age, coastline distance, and deck area.
Local optimization is performed via mini-batch stochastic gradient descent on the CTMC
log-likelihood, and only a 12-dimensional pseudo-gradient vector is uploaded to a central server
per communication round.
The server aggregates User updates using sample-weighted Federated Averaging (FedAvg) with
momentum and gradient clipping.
All experiments in this paper are conducted on fully synthetic data generated from a
known ground-truth parameter set with region-specific heterogeneity, enabling controlled
evaluation of federated convergence behaviour.
Simulation results across heterogeneous Users show consistent convergence of the average negative log-likelihood, with the aggregated gradient norm decreasing as User scale increases.
Furthermore, the federated update mechanism provides a natural participation incentive:
Users who register their local inspection datasets on a shared technical-standard platform
receive in return the periodically updated global benchmark parameters---information
that cannot be obtained from local data alone---thereby enabling evidence-based
life-cycle planning without surrendering data sovereignty.
\end{abstract}


\section{Introduction}

Bridge infrastructure underpins social and economic activity, yet managing large bridge portfolios
across thousands of municipalities remains a fundamental challenge in many countries.
In Japan, the 2014 amendment to the Road Act established a mandatory 5-year periodic inspection
cycle~\cite{mlit2019}, generating standardised damage-state records for each structural member.
These records capture transition information between damage states---information that is directly
applicable to Continuous-Time Markov Chain (CTMC) deterioration modelling.

Despite the richness of this data source, a critical obstacle prevents its full utilisation:
each municipality retains jurisdiction over its own inspection records, and sharing raw data across
organisational boundaries raises both privacy and legal governance concerns.
As a result, calibrating a shared national or regional benchmark deterioration model requires a
data federation strategy that preserves local ownership.

Continuous-Time Markov Chain (CTMC) hazard models have a long history in infrastructure
deterioration modelling~\cite{mauch2001,morcous2006,kalbfleisch1985}.
They provide a principled probabilistic framework for modelling state-to-state transitions observed
at discrete inspection intervals, and log-linear parameterisation naturally accommodates
covariate effects such as age, environmental exposure, and structural size.
However, fitting such models to decentralised data held by hundreds or thousands of municipalities
is not straightforward.

Federated Learning (FL), introduced by McMahan et al.~\cite{mcmahan2017}, offers a conceptually
clean solution: Users perform local optimisation and share only gradient information with a
central server, which aggregates updates and broadcasts an improved global model.
The Federated Averaging (FedAvg) algorithm~\cite{mcmahan2017} is the canonical approach,
and its convergence properties under partial participation and data heterogeneity have been
extensively analysed in subsequent work~\cite{li2020federated}.

Applications of FL to infrastructure health monitoring are emerging~\cite{li2022federated}, yet
its integration with CTMC-based hazard models---and specifically with the CTMC log-likelihood
gradient---has not been formally demonstrated.
This paper bridges that gap with four contributions:

\begin{enumerate}
  \item We formulate CTMC hazard model estimation as a federated optimisation problem and
        derive the mini-batch gradient of the CTMC log-likelihood suitable for local SGD.
  \item We implement a complete FedAvg pipeline (FedAvgClient / FedAvgServer) in PyTorch with
        partial participation, momentum, and gradient clipping.
  \item We design a synthetic data generation framework that faithfully reproduces
        the heterogeneous deterioration environments of coastal, riverside, and inland
        municipalities---enabling rigorous, reproducible evaluation.
  \item We conduct scale experiments across 500, 2{,}000, and 4{,}000 Users and characterise
        convergence behaviour, gradient stability, and communication cost.
\end{enumerate}

All experiments use \textbf{synthetic data only}.
No real inspection records are used; the synthetic generator is provided so that the methodology
can be validated and replicated without requiring access to restricted public-asset datasets.

The remainder of the paper is organised as follows.
Section~\ref{sec:related} reviews related work.
Section~\ref{sec:problem} formulates the CTMC hazard model and its log-likelihood.
Section~\ref{sec:method} describes the federated learning design.
Section~\ref{sec:experiments} presents the experimental setup and results.
Section~\ref{sec:discussion} discusses implications and limitations.
Section~\ref{sec:conclusion} concludes with directions for future work.

\section{Related Work}
\label{sec:related}

\subsection{Bridge Deterioration Modelling}

Markov-based models are the dominant paradigm for bridge and pavement deterioration
assessment~\cite{morcous2006,frangopol2004}.
Mauch and Madanat~\cite{mauch2001} introduced semi-parametric hazard rate models for bridge deck
deterioration and showed that CTMC formulations outperform discrete-time Markov chains when
inspection intervals are irregular.
Kalbfleisch and Lawless~\cite{kalbfleisch1985} provided the foundational panel-data likelihood for
CTMCs observed at arbitrary time points, which is the basis for our log-likelihood formulation.
Log-linear hazard functions naturally incorporate covariate effects via the exponential link,
ensuring hazard non-negativity without constrained optimisation.

Machine learning approaches have increasingly been applied to bridge deterioration and
condition assessment.
Huang~\cite{huang2010ann} demonstrated that artificial neural networks can predict
bridge condition ratings from inspection records, capturing non-linear covariate effects
beyond the reach of classical regression.
Bektas et al.~\cite{bektas2013tree} used classification trees to forecast NBI condition ratings
at scale, showing that ensemble methods substantially reduce prediction error relative to
ordinary least squares.
Cha et al.~\cite{cha2017deep} applied convolutional neural networks to crack detection in
concrete inspection images, extending data-driven assessment from tabular inspection scores
to raw visual data.
While these approaches demonstrate strong predictive accuracy, they do not yield
statistically interpretable transition rates for life-cycle planning.
In contrast, the present paper focuses on federated learning of parameters in a
statistical model that maximises a likelihood function, enabling inference on
physically meaningful hazard rates while preserving the data-privacy guarantees of a
federated framework.

\subsection{Federated Learning}

McMahan et al.~\cite{mcmahan2017} proposed FedAvg and demonstrated its effectiveness on
large-scale non-IID image classification.
Li et al.~\cite{li2020federated} proved convergence guarantees for FedAvg under partial
participation and bounded gradient heterogeneity.
Karimireddy et al.~\cite{karimireddy2020scaffold} introduced SCAFFOLD to correct for User
drift---an extension relevant to highly heterogeneous infrastructure data.
Bonawitz et al.~\cite{bonawitz2017} developed Secure Aggregation, which would further protect
uploaded gradient vectors in production deployments.

Beyond computer vision, FL has gained notable traction in the pharmaceutical and biomedical
domains, where strict regulatory constraints and commercial confidentiality prevent the
centralisation of sensitive experimental data.
Rieke et al.~\cite{rieke2020health} reviewed the role of FL in digital health and highlighted
drug-response modelling as a prime use case in which model parameters---rather than raw
molecular assay results---are shared among institutions.
The MELLODDY consortium demonstrated industry-scale FL across ten major pharmaceutical
companies: each partner retained proprietary compound--activity data on-premise, while a
coordinated FedAvg protocol learned a shared multi-task model for activity prediction,
yielding improvements in predictive accuracy that no single company could achieve
alone~\cite{oldenhof2023melloddy}.
Warnat-Herresthal et al.~\cite{warnat2021swarm} introduced Swarm Learning, a blockchain-based
variant of FL applied to transcriptomic disease classification, and showed that
decentralised parameter merging matches or exceeds centralised training even when local
datasets are small and heterogeneous.
These precedents from pharmaceutics attest that the value-exchange model underlying
federated parameter sharing---proprietary data remain local; a calibrated global model is
returned as the incentive---is viable at industrial scale, and motivate analogous
applications in infrastructure asset management.

\subsection{FL for Infrastructure and Structural Health Monitoring}

\paragraph{Deterioration Prediction.}
Markov-chain-based deterioration prediction has a long history in bridge asset management,
with landmark studies by Mauch and Madanat~\cite{mauch2001} and Morcous~\cite{morcous2006}
establishing log-linear hazard parameterisation as a robust choice for inspection-interval data.
More recently, computer-vision and machine-learning approaches have been applied to structural
health monitoring (SHM) for visual damage detection and condition scoring~\cite{spencer2019advances}.
Yet federated approaches to \emph{model-level} deterioration estimation---where a shared
statistical model is collaboratively calibrated across data-holding organisations---remain
largely under-explored.
Li et al.~\cite{li2022federated} surveyed FL for infrastructure health monitoring and
identified bridge and road deterioration as a high-priority application, noting the scarcity
of concrete implementations with domain-appropriate statistical models.

\paragraph{Technical Standard Platforms.}
National bridge management systems such as PONTIS (USA)~\cite{thompson1998pontis} and
Japan's Road Administration Information Management System~\cite{mlit2019} encode inspection
data in standardised formats and provide Markov-based performance prediction as built-in modules.
These platforms operate at the individual asset-owner level, however, and do not support
collaborative cross-organisation model calibration.
A federated extension of such platforms would allow municipalities and highway agencies to
contribute local deterioration observations without surrendering data sovereignty, enabling a
shared national benchmark while preserving the data governance constraints imposed by existing
law and policy.

\paragraph{FL Benchmarking Platforms.}
Reproducible evaluation is a prerequisite for adoption in engineering practice.
General-purpose FL benchmarking frameworks---LEAF~\cite{caldas2019leaf},
Flower~\cite{beutel2022flower}, and FedML~\cite{he2020fedml}---provide standardised
experimental protocols and heterogeneous data partitioning for common machine-learning tasks.
However, none of these frameworks includes survival-analysis or Markov-hazard benchmarks
appropriate for infrastructure applications.
Our work directly addresses this gap by providing a full CTMC-FedAvg implementation together
with a reproducible synthetic benchmark spanning 500 to 4{,}000 heterogeneous Users.

\section{Problem Formulation}
\label{sec:problem}

\subsection{Deterioration State Space and Transitions}

We model bridge members as evolving through three discrete deterioration states:

\begin{center}
\begin{tabular}{cl p{4.2cm}}
\toprule
State & Label & Description \\
\midrule
0 & Good   & No visible deterioration \\
1 & Minor  & Minor defects (damage class a/b) \\
2 & Severe & Requires repair (damage class c), absorbing \\
\bottomrule
\end{tabular}
\end{center}

Physical constraints restrict transitions to deterioration-direction only (no recovery):
\begin{equation}
  \mathcal{T} = \{(0\to1),\; (0\to2),\; (1\to2)\}
  \label{eq:transitions}
\end{equation}
State 2 is treated as absorbing ($\lambda_{2j} = 0$ for all $j$).

\subsection{Log-Linear Hazard Function}

Each active transition $(i\to j)\in\mathcal{T}$ is assigned a log-linear hazard:
\begin{equation}
  \lambda_{ij}(\mathbf{z}) = \exp\!\bigl(\beta_{0,ij} + \beta_{1,ij}z_1
    + \beta_{2,ij}z_2 + \beta_{3,ij}z_3\bigr)
  \label{eq:hazard}
\end{equation}
where $\mathbf{z} = (z_1, z_2, z_3)^{\top}$ is the covariate vector for the bridge member:

\begin{center}
\begin{tabular}{cl p{4.2cm}}
\toprule
Symbol & Covariate & Description \\
\midrule
$z_1$ & age      & Years since construction (normalised) \\
$z_2$ & sea dist & Distance from coastline, km (normalised) \\
$z_3$ & area     & Bridge deck area, m$^2$ (normalised) \\
\bottomrule
\end{tabular}
\end{center}

Normalisation uses each User's local maximum (local-max strategy).
The total hazard out of state $i$ is
$\Lambda_i(\mathbf{z}) = \sum_{j:(i\to j)\in\mathcal{T}} \lambda_{ij}(\mathbf{z})$.

\subsection{Transition Probabilities over Inspection Interval}

Following the CTMC panel-data framework~\cite{kalbfleisch1985}, for an inspection interval
$\Delta t$ the transition probabilities are:

\textbf{Stay probability} (no deterioration):
\begin{equation}
  p(i\to i\mid\mathbf{z},\Delta t)
  = \exp\!\bigl(-\Lambda_i(\mathbf{z})\,\Delta t\bigr)
  \label{eq:stay}
\end{equation}

\textbf{Move probability} (allowed transition $i\to j$, $j\neq i$):
\begin{equation}
  p(i\to j\mid\mathbf{z},\Delta t)
  = \frac{\lambda_{ij}(\mathbf{z})}{\Lambda_i(\mathbf{z})}
    \Bigl(1 - \exp\!\bigl(-\Lambda_i(\mathbf{z})\,\Delta t\bigr)\Bigr)
  \label{eq:move}
\end{equation}

\subsection{Log-Likelihood per Observation Pair}

For an observed consecutive-inspection pair
$(s_t = i,\; s_{t+1} = k,\; \Delta t,\; \mathbf{z})$:

\textbf{Case 1 ---No deterioration} ($k = i$):
\begin{equation}
  \ell_1 = -\Lambda_i(\mathbf{z})\,\Delta t
  \label{eq:ll_stay}
\end{equation}

\textbf{Case 2 ---Deterioration} ($k \neq i$, $(i\to k)\in\mathcal{T}$):
\begin{equation}
  \ell_2 = \log\lambda_{ik}(\mathbf{z})
         - \log\Lambda_i(\mathbf{z})
         + \log\!\Bigl(1 - \exp\!\bigl(-\Lambda_i(\mathbf{z})\,\Delta t\bigr)\Bigr)
  \label{eq:ll_move}
\end{equation}

The total User log-likelihood is $\mathcal{L}^u(\boldsymbol{\beta})
= \sum_{n}\ell_n$ over all observed pairs.
PyTorch automatic differentiation computes $\nabla_{\boldsymbol{\beta}}\mathcal{L}^u$
without hand-derivation of the Jacobian.

\subsection{Parameter Space}

The full parameter vector is
$\boldsymbol{\beta} \in \mathbb{R}^{12}$,
obtained by flattening the $3\times4$ coefficient matrix
(3 transitions $\times$ 4 coefficients per transition):

\begin{equation}
  \boldsymbol{\beta} =
  \bigl[\boldsymbol{\beta}_{0\to1},\;
        \boldsymbol{\beta}_{0\to2},\;
        \boldsymbol{\beta}_{1\to2}\bigr]^{\top},\quad
  \boldsymbol{\beta}_{ij} = (\beta_0,\beta_1,\beta_2,\beta_3)_{ij}
\end{equation}

This 12-dimensional vector is the \emph{only} quantity communicated between each User and
the server in every round.

\section{Federated Learning Design}
\label{sec:method}

\subsection{FedAvg with Partial Participation}

\begin{figure}[H]
\centering
\scalebox{1}[0.5]{\includegraphics[width=0.75\columnwidth]{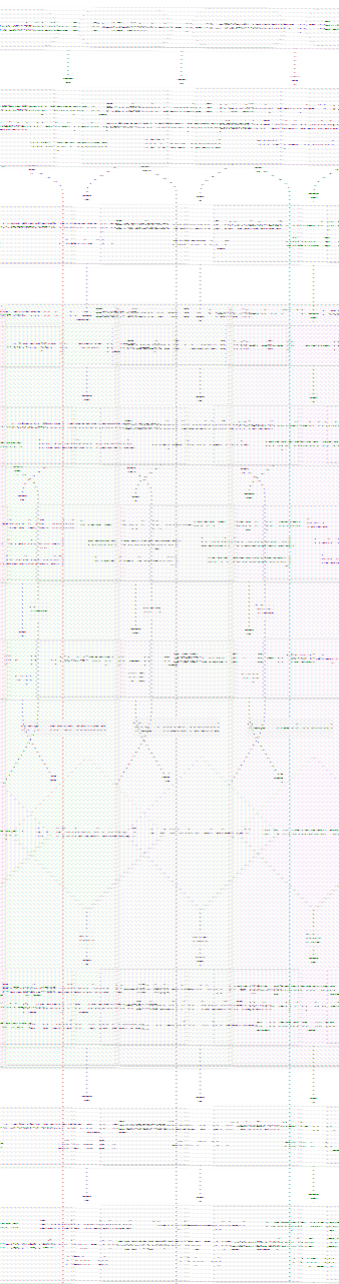}}
\caption{Computation flow of the proposed FedAvg-CTMC framework for bridge
  deterioration benchmark estimation.}
\label{fig:fedavg_flow}
\end{figure}

We adopt the Federated Averaging algorithm~\cite{mcmahan2017} extended with partial
User participation.  The end-to-end computation flow is illustrated in
Figure~\ref{fig:fedavg_flow}.
The numbered annotations correspond to each stage described below.

\textbf{Stage 1 ---Initialisation.}
The server initialises the global parameter vector
$\boldsymbol{\beta}^{(0)} = \mathbf{0}_{12}$ and the momentum buffer
$\mathbf{v}_0 = \mathbf{0}_{12}$.

\textbf{Stage 2 ---Partial participation and broadcast.}
At the start of round $r$, the server randomly samples a fraction $\rho$ of all
registered Users to form the active set $\mathcal{S}_r \subseteq \mathcal{U}$
and broadcasts the current $\boldsymbol{\beta}^{(r)}$ to each selected User.
Sampling a fresh subset every round reduces per-round communication cost and
models the practical reality that not every municipality responds to every data call.

\textbf{Stage 3 ---Local SGD with $K$ steps.}
Each active User $u\in\mathcal{S}_r$ saves
$\boldsymbol{\beta}_{\text{init}} \leftarrow \boldsymbol{\beta}^{(r)}$ and performs
$K$ mini-batch SGD steps on its local CTMC negative log-likelihood:
\begin{equation}
  \boldsymbol{\beta}^{(k+1)}_u
  = \boldsymbol{\beta}^{(k)}_u
    - \eta_{\text{local}}\,\nabla_{\boldsymbol{\beta}}
      \widehat{\mathcal{L}}^u(\boldsymbol{\beta}^{(k)}_u),
  \quad k = 0,\ldots,K-1
\end{equation}
where $\widehat{\mathcal{L}}^u$ is the mini-batch NLL computed from local
transition pairs $(s_t,s_{t+1},\Delta t,\mathbf{z})$.
All computation and data remain on the User device throughout this stage.

\textbf{Stage 4 ---Pseudo-gradient upload.}
After $K$ steps, User $u$ computes the pseudo-gradient
\begin{equation}
  \mathbf{g}_u
  = \frac{\boldsymbol{\beta}_{\text{init}} - \boldsymbol{\beta}^{(K)}_u}{\eta_{\text{local}}}
  \label{eq:pseudo_grad}
\end{equation}
and uploads the pair $(\mathbf{g}_u, n_u)$ to the server---a payload of only $\approx$52 bytes.
No inspection record, bridge identifier, or intermediate activation is transmitted.

\textbf{Stage 5 ---Sample-weighted aggregation and server update.}
The server gathers all uploaded pairs and computes the weighted average gradient:
\begin{equation}
  \bar{\mathbf{g}}_r
  = \frac{\sum_{u\in\mathcal{S}_r} n_u\,\mathbf{g}_u}{\sum_{u\in\mathcal{S}_r} n_u}
  \label{eq:aggregation}
\end{equation}
It then applies $\ell_2$ gradient clipping (threshold $\delta=1.0$) and
momentum ($\mu = 0.9$) before updating the global model:
\begin{align}
  \mathbf{v}_r &= \mu\,\mathbf{v}_{r-1} + \text{clip}(\bar{\mathbf{g}}_r,\delta) \\
  \boldsymbol{\beta}^{(r+1)} &= \boldsymbol{\beta}^{(r)} - \eta_{\text{global}}\,\mathbf{v}_r
  \label{eq:server_update}
\end{align}
The updated $\boldsymbol{\beta}^{(r+1)}$ becomes the broadcast value for round $r+1$,
closing the federated loop.

The complete procedure is summarised in Algorithm~\ref{alg:fedavg}.

\begin{algorithm}[t]
\caption{FedAvg-CTMC for Bridge Deterioration}
\label{alg:fedavg}
\begin{algorithmic}[1]
\REQUIRE Users $\mathcal{U}$, rounds $R$, fraction $\rho$, local steps $K$,
         $\eta_{\text{local}}$, $\eta_{\text{global}}$, momentum $\mu$
\STATE Initialise $\boldsymbol{\beta}^{(0)} \leftarrow \mathbf{0}_{12}$,\;
       $\mathbf{v}_0 \leftarrow \mathbf{0}_{12}$
\FOR{$r = 1, \ldots, R$}
  \STATE Sample $\mathcal{S}_r \subset \mathcal{U}$,\; $|\mathcal{S}_r|=\lceil\rho|\mathcal{U}|\rceil$
  \FOR{each User $u \in \mathcal{S}_r$}
    \STATE Receive $\boldsymbol{\beta}^{(r)}$;\;
           set $\boldsymbol{\beta}_u^{(0)} \leftarrow \boldsymbol{\beta}^{(r)}$
    \FOR{$k = 0, \ldots, K-1$}
      \STATE Draw mini-batch $\mathcal{B}_k$ from local pairs
      \STATE $\boldsymbol{\beta}_u^{(k+1)} \leftarrow \boldsymbol{\beta}_u^{(k)}
             - \eta_{\text{local}}\nabla_{\boldsymbol{\beta}}
               \widehat{\mathcal{L}}^u(\boldsymbol{\beta}_u^{(k)};\mathcal{B}_k)$
    \ENDFOR
    \STATE Compute $\mathbf{g}_u \leftarrow
           (\boldsymbol{\beta}^{(r)} - \boldsymbol{\beta}_u^{(K)})/\eta_{\text{local}}$
    \STATE Upload $(\mathbf{g}_u,\, n_u)$ to server
  \ENDFOR
  \STATE $\bar{\mathbf{g}}_r \leftarrow
         \sum_u n_u\mathbf{g}_u \;/\; \sum_u n_u$
  \STATE $\mathbf{v}_r \leftarrow \mu\mathbf{v}_{r-1}
         + \text{clip}(\bar{\mathbf{g}}_r,\,\delta=1.0)$
  \STATE $\boldsymbol{\beta}^{(r+1)} \leftarrow
         \boldsymbol{\beta}^{(r)} - \eta_{\text{global}}\,\mathbf{v}_r$
\ENDFOR
\RETURN $\boldsymbol{\beta}^{(R)}$
\end{algorithmic}
\end{algorithm}

\subsection{Communication Cost}

The 12-dimensional parameter/gradient representation keeps communication overhead minimal:

\begin{center}
\begin{tabular}{p{3.6cm}p{3.4cm}}
\toprule
Item & Size \\
\midrule
Global model broadcast & $12\times4$ bytes $= 48$ bytes \\
User gradient upload & $12\times4$ bytes $+$ sample count $\approx 52$ bytes \\
Per-round server traffic (400 Users) & $<25$ KB \\
\bottomrule
\end{tabular}
\end{center}

This negligible bandwidth usage makes the framework viable even over low-bandwidth municipal
network connections.

\begin{figure*}[t]
\centering
\hfil\includegraphics[width=0.97\textwidth]{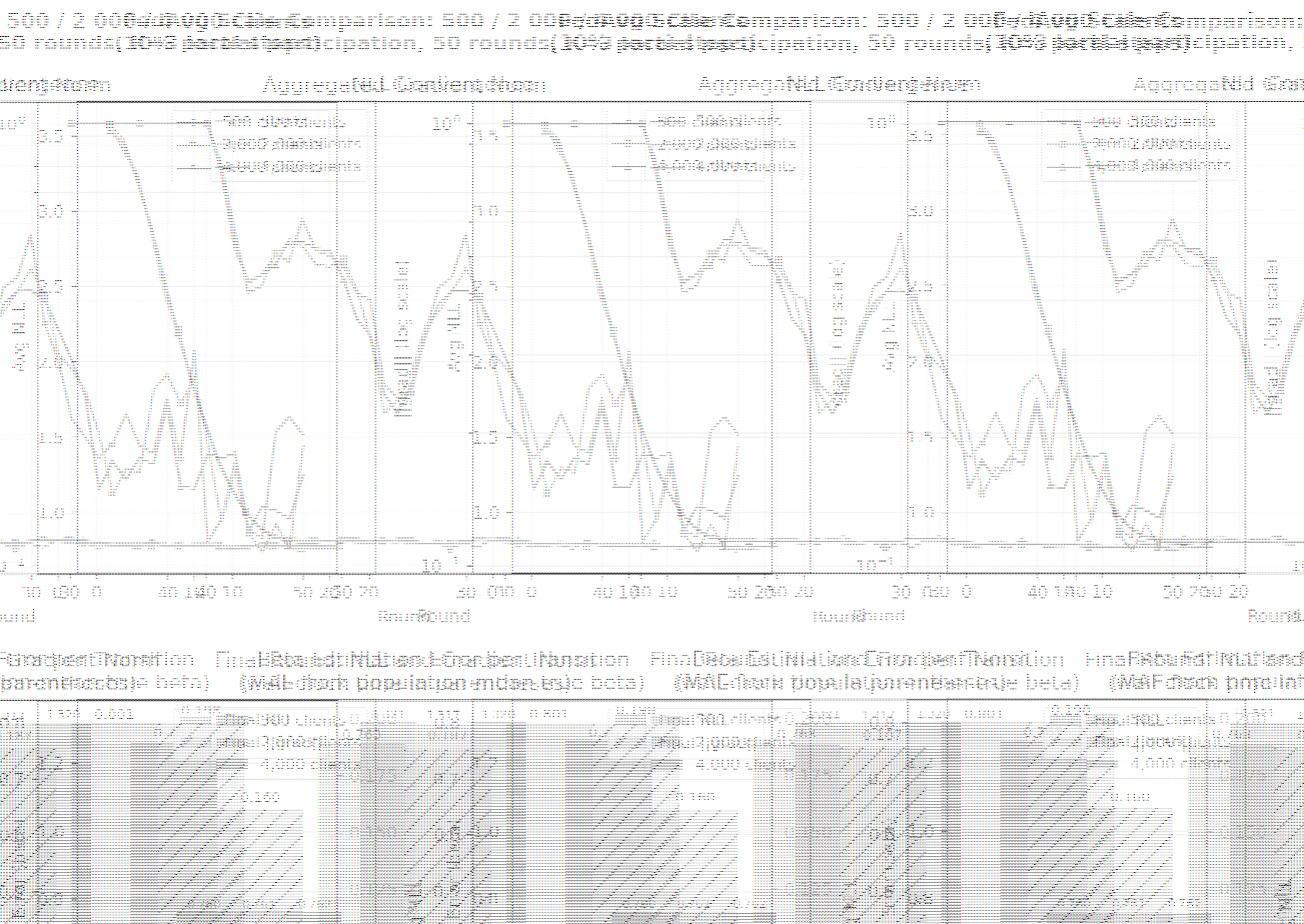}\hfil
\caption{Four-panel scale comparison for 500, 2{,}000, and 4{,}000 Users.
  \textbf{Top-left}: NLL convergence curves.
  \textbf{Top-right}: Aggregated gradient norm in log scale.
  \textbf{Bottom-left}: Beta estimation MAE per transition (grouped bar).
  \textbf{Bottom-right}: Final-round NLL and $\|\bar{\mathbf{g}}\|$ with wall-clock time.}
\label{fig:scale}
\end{figure*}

\section{Experiments}
\label{sec:experiments}

\subsection{Synthetic Data Generation}
\label{sec:data}

\textbf{All experiments use fully synthetic data.}
No real bridge inspection records were used in this study.
A ground-truth parameter matrix $\boldsymbol{\beta}^{*}\in\mathbb{R}^{3\times4}$ is fixed:

\begin{center}
\begin{tabular}{lrrrr}
\toprule
Transition & $\beta_0$ & $\beta_1$ & $\beta_2$ & $\beta_3$ \\
\midrule
$0\to1$ & $-2.0$ & $+0.5$ & $-0.3$ & $+0.10$ \\
$0\to2$ & $-4.0$ & $+0.3$ & $-0.5$ & $+0.05$ \\
$1\to2$ & $-2.5$ & $+0.4$ & $-0.4$ & $+0.08$ \\
\bottomrule
\end{tabular}
\end{center}

Each synthetic municipality is assigned a \emph{region type} that determines its coastline
distance range, bridge count distribution, and local $\boldsymbol{\beta}$ noise level:

\begin{center}
\begin{tabular}{lc p{2cm} c}
\toprule
Region & Proportion & Bridge count & $\beta$ noise $\sigma$ \\
\midrule
Coastal   & 30\% & log-normal $(10$--$80)$  & 0.20 \\
Riverside & 30\% & log-normal $(5$--$50)$   & 0.15 \\
Inland    & 40\% & log-normal $(3$--$30)$   & 0.10 \\
\bottomrule
\end{tabular}
\end{center}

Each User's local true $\boldsymbol{\beta}$ is obtained by adding zero-mean Gaussian noise
$\mathcal{N}(\mathbf{0},\sigma^2\mathbf{I})$ to $\boldsymbol{\beta}^{*}$, simulating
heterogeneous deterioration environments.
Member count per bridge is drawn uniformly from $\{1,2,3\}$; inspection count per member
from $\{2,3,4,5\}$.
Transition pairs $(s_t, s_{t+1}, \Delta t, \mathbf{z})$ are extracted from each User's
simulated inspection record and used as the local dataset.

\subsection{Hyperparameters}

All scale experiments use the following fixed hyperparameters:

\begin{center}
\begin{tabular}{p{4.5cm}p{2.5cm}}
\toprule
Hyperparameter & Value \\
\midrule
Number of rounds $R$ & 50 \\
User fraction $\rho$ & 10\% \\
Local steps $K$ & 3 \\
Local learning rate $\eta_{\text{local}}$ & 0.01 \\
Global learning rate $\eta_{\text{global}}$ & 0.05 \\
Mini-batch size & 32 \\
Momentum $\mu$ & 0.9 \\
Gradient clip norm $\delta$ & 1.0 \\
Random seed & 2024 \\
\bottomrule
\end{tabular}
\end{center}

\subsection{Scale Comparison: Heterogeneous Users}

Table~\ref{tab:scale} summarises the scale-up results.
As User count increases from 500 to 4{,}000, the per-round sample size grows proportionally
and the aggregated gradient norm $\|\bar{\mathbf{g}}_R\|$ decreases monotonically
(0.354 $\to$ 0.140), indicating more stable gradient estimation with larger participation.
The final NLL consistently converges to $\approx$0.76--0.80 across all scales within 50 rounds,
demonstrating robustness to heterogeneity.

\begin{table}[H]
\centering
\caption{Scale comparison results (50 rounds, 10\% participation rate).}
\label{tab:scale}
\begin{tabular}{lrrr}
\toprule
Metric & 500 & 2{,}000 & 4{,}000 \\
\midrule
Transition pairs (total) & 70{,}487 & 278{,}627 & 552{,}844 \\
Users per round        & 50       & 200       & 400       \\
Avg samples/round        & $\sim$6{,}000 & $\sim$28{,}000 & $\sim$55{,}000 \\
Round~1 Avg NLL          & 3.42     & 3.57      & 3.55      \\
Round~50 Avg NLL         & 0.775    & 0.763     & 0.766     \\
Round~50 $\|\bar{\mathbf{g}}\|$ & 0.354 & 0.175 & 0.140 \\
Wall-clock time (s)      & 19       & 73        & 171       \\
\bottomrule
\end{tabular}
\end{table}

Figure~\ref{fig:scale} visualises the four convergence metrics simultaneously.

\textbf{Top-left: NLL convergence curves.}
All three scales exhibit a rapid NLL drop within the first 10 rounds followed by a gradual
plateau, confirming that 50 rounds are sufficient for practical convergence.
The final NLL values (0.775, 0.763, 0.766 for 500, 2{,}000, and 4{,}000 Users) are
nearly identical, demonstrating that the FedAvg-CTMC objective is scale-invariant once
sufficient transition pairs are available.

\textbf{Top-right: Aggregated gradient norm (log scale).}
The monotone decrease of $\|\bar{\mathbf{g}}_r\|$ from round 1 to round 50 confirms
that the server update direction consistently reduces residual gradient magnitude.
Critically, the final-round norm decreases from 0.354 (500 Users) to 0.140 (4{,}000 Users),
indicating that larger participation provides a more precise gradient signal through
the sample-weighted aggregation in Eq.~(\ref{eq:aggregation}).
This reduction in gradient noise at larger scale suggests that the framework would
benefit from a higher User population in production deployments.

\textbf{Bottom-left: Beta estimation MAE per transition (grouped bar).}
The mean absolute error between the learned $\boldsymbol{\beta}$ and the ground-truth
$\boldsymbol{\beta}^{*}$ is shown grouped by transition and by scale.
The MAE values do not decrease substantially with scale, which is the expected behaviour:
because each User's local true $\boldsymbol{\beta}$ is itself a noisy perturbation of
$\boldsymbol{\beta}^{*}$, the federated optimum converges to the population-weighted mean
rather than the single ground-truth point.
The residual MAE therefore reflects irreducible population heterogeneity, not estimation
inefficiency.

\textbf{Bottom-right: Final-round NLL, gradient norm, and wall-clock time.}
The bar chart confirms the scalability trade-off: NLL and gradient norm are stable across
scales, while wall-clock time grows near-linearly (19~s $\to$ 73~s $\to$ 171~s).
In a real deployment, User-side computation would execute in parallel, so round latency
would be bounded by the slowest participating User rather than by the total User count,
making the framework practically feasible even at 4{,}000-User scale.

\subsection{Learning Curves (4{,}000-User Run)}

\begin{figure*}[t]
\centering
\includegraphics[width=0.9\textwidth]{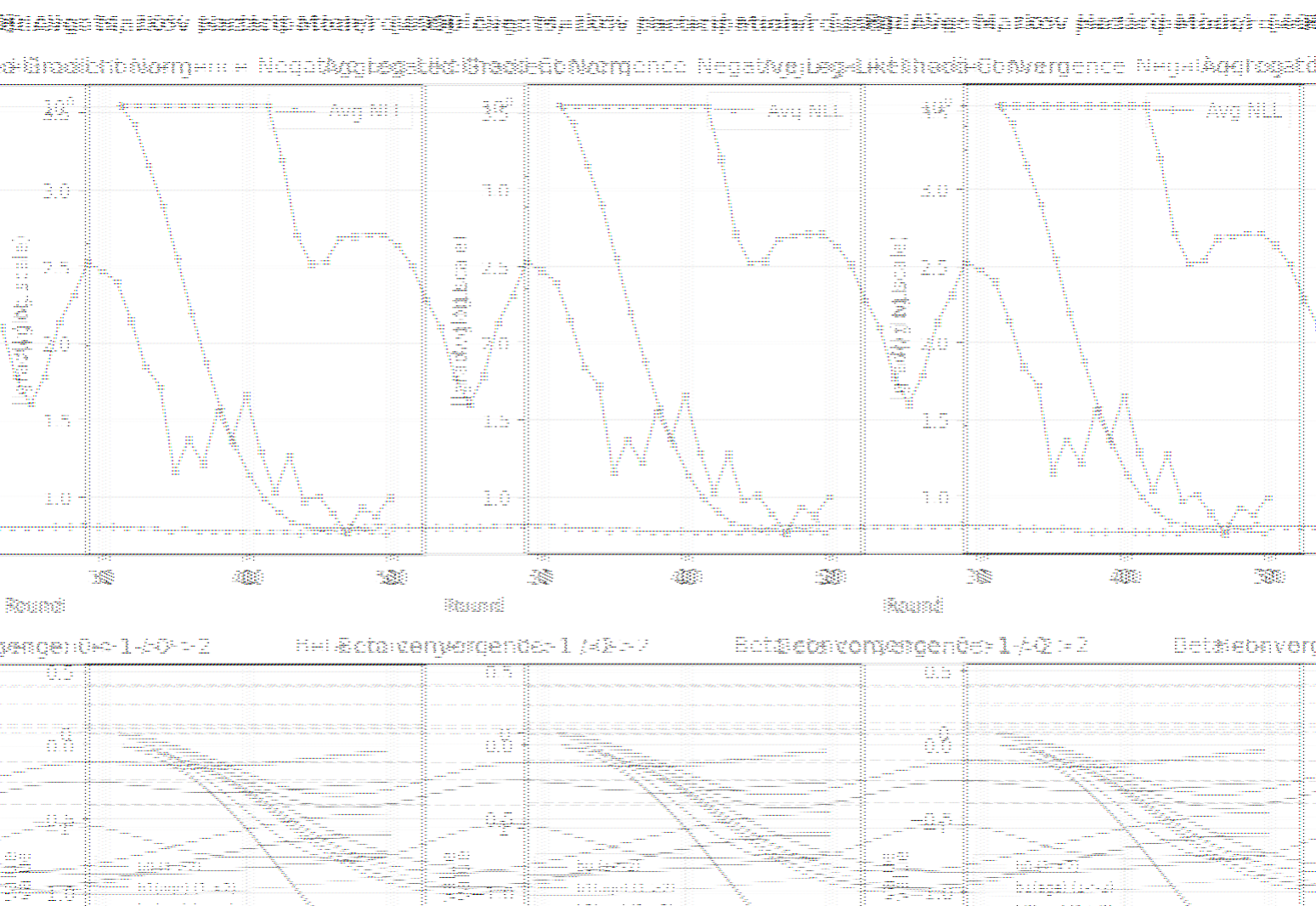}
\caption{Learning curves for the 4{,}000-User run (50 rounds, $\rho=10\%$).
  \textbf{Top-left}: Average NLL per communication round.
  \textbf{Top-right}: Aggregated gradient norm $\|\bar{\mathbf{g}}_r\|$ in log scale.
  \textbf{Bottom-left}: Trajectories of 8 $\beta$ coefficients for transitions
  $0\to1$ and $0\to2$; dashed lines show ground-truth values.
  \textbf{Bottom-right}: Trajectories of 4 $\beta$ coefficients for transition
  $1\to2$; dashed lines show ground-truth values.}
\label{fig:lc}
\end{figure*}

Figure~\ref{fig:lc} shows the intra-training dynamics for the 4{,}000-User experiment
in a four-panel layout.
The top-left panel plots the per-round average NLL: after an initial rapid drop in the
first 10 rounds, the curve flattens smoothly, confirming that 50 rounds are sufficient
for practical convergence under the chosen hyperparameters.
The top-right panel plots the aggregated gradient norm on a logarithmic scale;
its monotone decrease from $\approx$1.8 to 0.14 indicates that the server update is
consistently reducing the residual gradient magnitude without oscillation.
The bottom panels trace the 12 individual $\beta$ coefficients across rounds, grouped
by transition (0$\to$1/0$\to$2 in the left panel; 1$\to$2 in the right).
Dashed horizontal lines mark the ground-truth values used for data generation.
All coefficients converge to stable values within 30 rounds, and the systematic
offsets from the dashed lines reflect the population-average nature of the federated
objective discussed in Section~\ref{sec:discussion}.

\subsection{Learned Parameters and Beta Estimation Error}

Table~\ref{tab:beta} compares the ground-truth $\boldsymbol{\beta}^{*}$ with the learned global
$\boldsymbol{\beta}$ from the 4{,}000-User run.
The systematic positive offset in $\beta_0$ reflects the population-average over the heterogeneous
User distribution: each User's local true $\boldsymbol{\beta}$ deviates from
$\boldsymbol{\beta}^{*}$ by region-specific Gaussian noise, so the federated optimum
is a weighted average across Users rather than the single ground-truth value.
This is the intended behaviour for benchmark estimation, where the global model should
represent the \emph{population-level} deterioration environment rather than any individual User.

\begin{table}[H]
\centering
\caption{Learned vs.\ ground-truth $\boldsymbol{\beta}$ (4{,}000-User run, round 50).}
\label{tab:beta}
\begin{tabular}{llrrr}
\toprule
Transition & Coef & True & Learned & Error \\
\midrule
$0\to1$ & $\beta_0$ & $-2.000$ & $-1.273$ & $+0.727$ \\
        & $\beta_1$ & $+0.500$ & $-0.207$ & $-0.707$ \\
        & $\beta_2$ & $-0.300$ & $-0.526$ & $-0.226$ \\
        & $\beta_3$ & $+0.100$ & $-0.187$ & $-0.287$ \\
\midrule
$0\to2$ & $\beta_0$ & $-4.000$ & $-2.666$ & $+1.334$ \\
        & $\beta_1$ & $+0.300$ & $-1.302$ & $-1.602$ \\
        & $\beta_2$ & $-0.500$ & $-1.510$ & $-1.010$ \\
        & $\beta_3$ & $+0.050$ & $-1.283$ & $-1.333$ \\
\midrule
$1\to2$ & $\beta_0$ & $-2.500$ & $-1.212$ & $+1.288$ \\
        & $\beta_1$ & $+0.400$ & $-0.599$ & $-0.999$ \\
        & $\beta_2$ & $-0.400$ & $-0.617$ & $-0.217$ \\
        & $\beta_3$ & $+0.080$ & $-0.462$ & $-0.542$ \\
\bottomrule
\end{tabular}
\end{table}

\subsection{Transition Probabilities}

Table~\ref{tab:prob} reports the global model's predicted transition probabilities for three
representative covariate scenarios with $\Delta t = 3$ years.

\begin{table*}[t]
\centering
\caption{Transition+ probabilities from the global model
  (4{,}000-User run, $\Delta t = 3$~yr).}
\label{tab:prob}
\begin{tabular}{p{4.2cm}p{5.8cm}p{3.6cm}}
\toprule
Scenario ($z_1,z_2,z_3$) & State 0 & State 1 \\
\midrule
Young, far, small $(0.2,0.8,0.1)$ &
  ${\to}0$: 0.570, ${\to}1$: 0.398, ${\to}2$: 0.032 &
  ${\to}1$: 0.630, ${\to}2$: 0.370 \\
Mid-age, near, medium $(0.5,0.3,0.5)$ &
  ${\to}0$: 0.535, ${\to}1$: 0.438, ${\to}2$: 0.027 &
  ${\to}1$: 0.646, ${\to}2$: 0.354 \\
Old, near, large $(0.9,0.1,0.9)$ &
  ${\to}0$: 0.562, ${\to}1$: 0.425, ${\to}2$: 0.013 &
  ${\to}1$: 0.724, ${\to}2$: 0.276 \\
\bottomrule
\end{tabular}
\end{table*}

\begin{figure}[H]
\centering
\includegraphics[width=1.035\columnwidth]{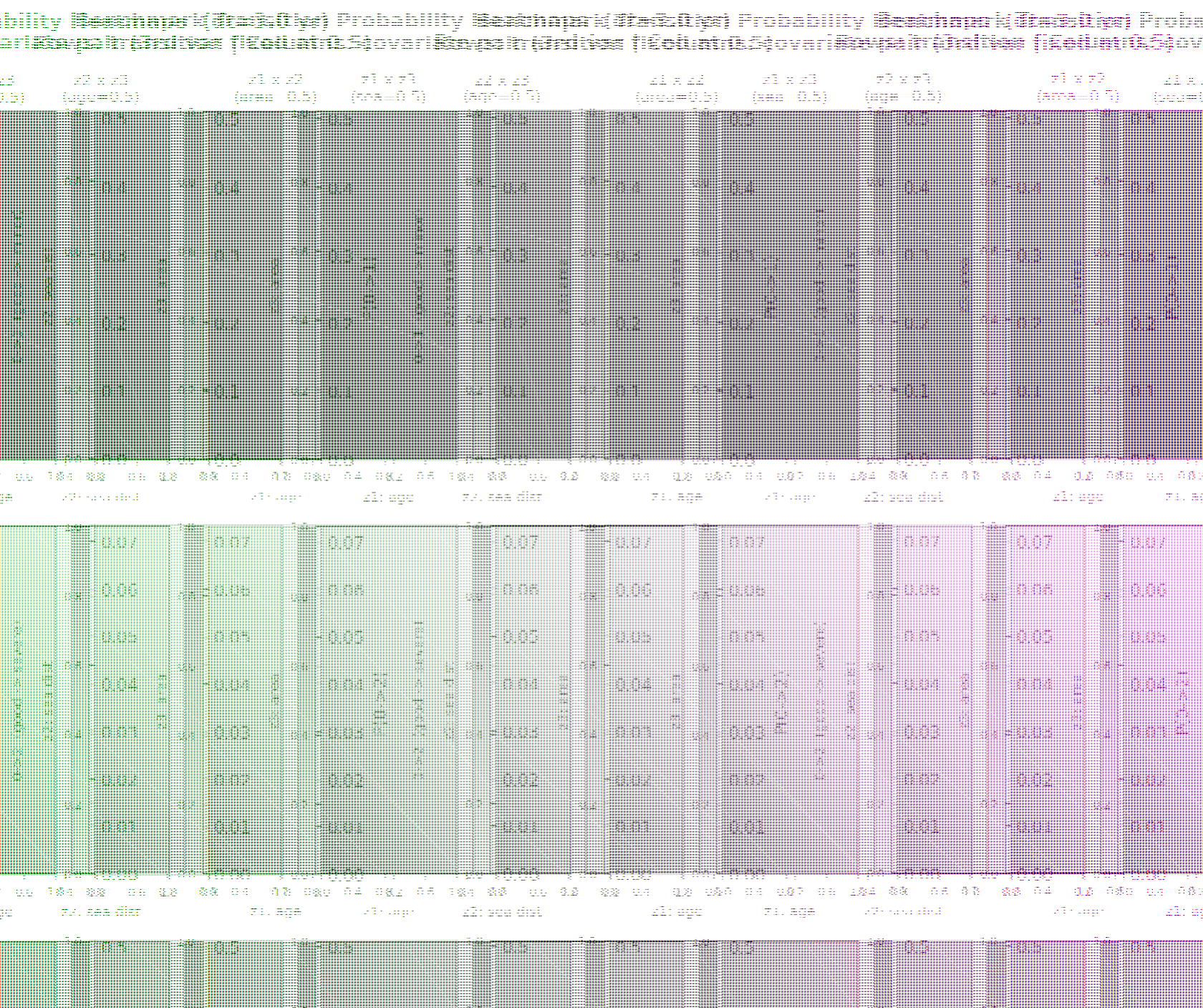}
\caption{3$\times$3 covariate-pair heatmaps of benchmark transition probabilities
  ($\Delta t = 3$~yr, 4{,}000-User model).
  \textbf{Rows}: transitions $0\to1$ / $0\to2$ / $1\to2$.
  \textbf{Columns}: covariate pairs $(z_1,z_2)$ / $(z_1,z_3)$ / $(z_2,z_3)$;
  the remaining covariate is fixed at 0.5.
  Colour scale is shared within each row; white contour lines mark iso-probability levels.}
\label{fig:heatmap}
\end{figure}

Figure~\ref{fig:heatmap} displays transition probability heatmaps over covariate pairs.
The $0\to1$ transition probability increases with bridge age ($z_1$) and is attenuated by greater
sea distance ($z_2$), consistent with the positive $\beta_1$ and negative $\beta_2$ in the
ground-truth parameters.
The $0\to2$ transition remains low across all scenarios, reflecting the large negative
$\beta_0 = -4.0$ that suppresses direct good-to-severe transitions.

\section{Discussion}
\label{sec:discussion}

\textbf{Limitations of synthetic evaluation.}
The synthetic data generator assumes independence between bridges and a fixed three-state model.
Real inspection data exhibit correlated members within a bridge, irregular observation plans,
interval-censored transitions, and state assessment errors---all of which require extensions
beyond the present formulation.
We regard this work as a feasibility demonstration; validation on real data is reserved for
future research under an appropriate data governance agreement.

\textbf{Participation incentive through benchmark access.}
A practical question for real-world deployment is why individual Users would volunteer their
inspection records to a federated platform.
We propose a value-exchange mechanism grounded in the asymmetry between local and global
estimation capacity.
Each User registers their inspection dataset on a shared technical-standard platform managed
by a neutral operator desk and receives a
locally fitted $\hat{\boldsymbol{\beta}}^{(k)}$ in return.
After each aggregation round, the operator redistributes the updated global model
$\boldsymbol{\beta}^{(t)}$ to all participating Users.
Because the number of observed state transitions within a single municipality is typically
too small to identify the full covariate structure of the hazard model, the global benchmark
provides information that is unattainable from local data alone.
In particular, Users gain access to national-average transition rate estimates that can serve
as empirical priors for life-cycle cost analysis, safety assessment, and budget planning.
This information asymmetry---local estimation is feasible but imprecise; global estimation
requires pooled data but yields reliable benchmarks---constitutes a natural incentive for
voluntary participation without requiring monetary transfers or regulatory mandates.

\section{Conclusion}
\label{sec:conclusion}

We presented a federated framework for estimating CTMC hazard models of bridge deterioration
using Federated Averaging.
The methodology combines a log-linear CTMC log-likelihood with PyTorch automatic differentiation
for gradient computation, and a sample-weighted FedAvg server with momentum and gradient clipping
for robust aggregation.

Experiments on fully synthetic data across 500, 2{,}000, and 4{,}000 heterogeneous Users
demonstrated consistent NLL convergence to $\approx$0.76--0.80 within 50 rounds, with gradient
norm decreasing from 0.354 to 0.140 as scale increases.
Communication overhead is negligible ($<$25~KB per round for 400 Users),
confirming practical feasibility.

Future directions include: (i) extension to multi-member bridge models with correlated
observations; (ii) integration of Secure Aggregation~\cite{bonawitz2017} for cryptographic
gradient privacy; (iii) incorporation of state assessment uncertainty via hidden Markov extensions;
and (iv) validation on real municipal inspection datasets.
The open-source implementation is available at
\url{https://github.com/tk-yasuno/markov_hazard_fedavg}.

\section*{Acknowledgements}

The author thanks colleagues at Yachiyo Engineering Co., Ltd.\ for discussions on bridge
inspection practice and data governance.
All numerical experiments were conducted on a personal workstation;
no confidential data were ever used in this study.

\bibliographystyle{unsrt}
\bibliography{fedavg_ctmc_bridge_2026}

\begin{thebibliography}{10}

\bibitem{mlit2019}
{Ministry of Land, Infrastructure, Transport and Tourism (MLIT)}.
\newblock Manual for bridge periodic inspection (doro-hashi teiki tenken
  yoryo).
\newblock Technical report, Road Bureau, MLIT, Japan, 2019.

\bibitem{mauch2001}
Matthew Mauch and Samer Madanat.
\newblock Semiparametric hazard rate models of reinforced concrete bridge deck
  deterioration.
\newblock {\em Journal of Infrastructure Systems}, 7(2):49--57, 2001.

\bibitem{morcous2006}
George Morcous.
\newblock Performance prediction of bridge deck systems using {M}arkov chains.
\newblock {\em Journal of Performance of Constructed Facilities},
  20(2):146--153, 2006.

\bibitem{kalbfleisch1985}
John~D. Kalbfleisch and Jerald~F. Lawless.
\newblock The analysis of panel data under a {M}arkov assumption.
\newblock {\em Journal of the American Statistical Association},
  80(392):863--871, 1985.

\bibitem{mcmahan2017}
H.~Brendan McMahan, Eider Moore, Daniel Ramage, Seth Hampson, and
  Blaise~Ag\"{u}era y~Arcas.
\newblock Communication-efficient learning of deep networks from decentralized
  data.
\newblock In {\em Proceedings of the 20th International Conference on
  Artificial Intelligence and Statistics (AISTATS)}, volume~54 of {\em PMLR},
  pages 1273--1282, 2017.

\bibitem{li2020federated}
Tian Li, Anit~Kumar Sahu, Manzil Zaheer, Maziar Sanjabi, Alexander Smola,
  et~al.
\newblock Federated optimization in heterogeneous networks.
\newblock {\em Proceedings of Machine Learning and Systems (MLSys)},
  2:429--450, 2020.

\bibitem{li2022federated}
Danilenkov Li, Arno Knobbe, H.~Jaap van~den Herik, and Sandjai Bhulai.
\newblock Federated learning for infrastructure health monitoring: A survey.
\newblock {\em arXiv preprint arXiv:2206.00009}, 2022.

\bibitem{frangopol2004}
Dan~M. Frangopol, Maarten-Jan Kallen, and Jan~M. van Noortwijk.
\newblock Probabilistic models for life-cycle performance of deteriorating
  structures: Review and future directions.
\newblock {\em Progress in Structural Engineering and Materials},
  6(4):197--212, 2004.

\bibitem{huang2010ann}
Ying-Hua Huang.
\newblock Artificial neural network model of bridge deterioration.
\newblock {\em Journal of Performance of Constructed Facilities},
  24(6):597--602, 2010.

\bibitem{bektas2013tree}
Basak~A. Bektas, Alicia Carriquiry, and Omar Smadi.
\newblock Using classification trees for predicting national bridge inventory
  condition ratings.
\newblock {\em Journal of Infrastructure Systems}, 19(4):425--433, 2013.

\bibitem{cha2017deep}
Young-Jin Cha, Wooram Choi, and Oral B\"uy\"uk\"ozt\"urk.
\newblock Deep learning-based crack damage detection using convolutional neural
  networks.
\newblock {\em Computer-Aided Civil and Infrastructure Engineering},
  32(5):361--378, 2017.

\bibitem{karimireddy2020scaffold}
Sai~Praneeth Karimireddy, Satyen Kale, Mehryar Mohri, Sashank~J. Reddi,
  Sebastian~U. Stich, et~al.
\newblock {SCAFFOLD}: Stochastic controlled averaging for federated learning.
\newblock In {\em Proceedings of the 37th International Conference on Machine
  Learning (ICML)}, volume 119 of {\em PMLR}, pages 5132--5143, 2020.

\bibitem{bonawitz2017}
Keith Bonawitz, Vladimir Ivanov, Ben Kreuter, Antonio Marcedone, H.~Brendan
  McMahan, et~al.
\newblock Practical secure aggregation for privacy-preserving machine learning.
\newblock In {\em Proceedings of the 2017 ACM SIGSAC Conference on Computer and
  Communications Security (CCS)}, pages 1175--1191, 2017.

\bibitem{rieke2020health}
Nicola Rieke, Jonny Hancox, Wenqi Li, Fausto Milletari, Holger~R. Roth,
  et~al.
\newblock The future of digital health with federated learning.
\newblock {\em npj Digital Medicine}, 3:119, 2020.

\bibitem{oldenhof2023melloddy}
Martijn Oldenhof, Antonius Hasselgren, Jaak Stroobants, Emmanuel Gustin, Yves
  Moreau, et~al.
\newblock Industry-scale orchestrated federated learning for drug discovery.
\newblock {\em arXiv preprint arXiv:2210.08871}, 2023.

\bibitem{warnat2021swarm}
Stefanie Warnat-Herresthal, Hartmut Schultze, K.~Lakshminarayan Shastry,
  Sathyanarayanan Manamohan, Saikat Mukherjee, et~al.
\newblock Swarm learning for decentralized and confidential clinical machine
  learning.
\newblock {\em Nature}, 594:265--270, 2021.

\bibitem{spencer2019advances}
Billie~F. Spencer, Vedhus Hoskere, and Yasutaka Narazaki.
\newblock Advances in computer vision-based civil infrastructure inspection and
  monitoring.
\newblock {\em Engineering}, 5(2):199--222, 2019.

\bibitem{thompson1998pontis}
Paul~D. Thompson, Eric~P. Small, Mark Johnson, and Andrew~R. Marshall.
\newblock The {Pontis} bridge management system.
\newblock {\em Structural Engineering International}, 8(4):303--308, 1998.

\bibitem{caldas2019leaf}
Sebastian Caldas, Peter Wu, Tian Li, Jakub Konecny, H.~Brendan McMahan,
  et~al.
\newblock {LEAF}: A benchmark for federated settings.
\newblock {\em arXiv preprint arXiv:1812.01097}, 2019.

\bibitem{beutel2022flower}
Daniel~J. Beutel, Taner Topal, Akhil Mathur, Xinchi Qiu, Javier
  Fernandez-Marques, et~al.
\newblock Flower: A friendly federated learning research framework.
\newblock {\em arXiv preprint arXiv:2007.14390}, 2022.

\bibitem{he2020fedml}
Chaoyang He, Songze Li, Jinhyun So, Mi~Zhang, Hongyi Wang, et~al.
\newblock {FedML}: A research library and benchmark for federated machine
  learning.
\newblock In {\em NeurIPS Workshop on Scalability, Privacy, and Security in
  Federated Learning}, 2020.

\end{thebibliography}

\end{document}